\documentclass[journal]{IEEEtran}
\usepackage{amsthm}
\usepackage{empheq}
\usepackage{cancel}
\usepackage{mdframed,xcolor}
\usepackage{tcolorbox}
\usepackage{placeins}
\usepackage{relsize}
\usepackage{setspace}
\usepackage{stmaryrd}
\usepackage{xcolor}
\usepackage{url}
\usepackage{multirow}
\usepackage{amssymb}
\usepackage{amsmath}
\usepackage{array}
\usepackage[end]{algpseudocode}
\usepackage{algorithm}
\usepackage{amsfonts}
\usepackage{diagbox}
\usepackage{booktabs}
\usepackage{filecontents}
\usepackage{mathtools}
\usepackage{caption}
\newcommand{\BS}{\mathrm{BS}}
\usepackage{subcaption}
\usepackage{enumitem}
\usepackage{xfrac}
\usepackage{nicefrac}
\usepackage{cite}
\usepackage{url}
\usepackage{soul}
\usepackage{algorithm}
\algdef{SE}{Begin}{End}{\textbf{begin}}{\textbf{end}}
\setlist[itemize]{leftmargin=*}
\usepackage{xcolor}
\definecolor{rv1}{rgb}{1.0, 0.44, 0.37}
\definecolor{rv2}{rgb}{0.4, 1.0, 0.0}
\definecolor{rv3}{rgb}{0.0, 0.75, 1.0}
\definecolor{rvt}{rgb}{0.75, 0.75, 0.75}
\definecolor{my-blue}{cmyk}{0.1, 0.0, 0.0, 0.0, 1.00}
\usepackage{soul}

\newtheoremstyle{exampstyle}
{7pt} % Space above
{7pt} % Space below
{\itshape} % Body font
{} % Indent amount
{\bfseries} % Theorem head font
{.} % Punctuation after theorem head
{.5em} % Space after theorem head
{} % Theorem head spec (can be left empty, meaning `normal')
\theoremstyle{exampstyle}

\begin{document}

\title{\huge Wireless TokenCom: RL-Based Tokenizer Agreement for Multi-User Wireless Token Communications   

%:\\ A hybrid RL Solution
}

\author{Farshad Zeinali,~\IEEEmembership{Student Member, IEEE}, Mahdi Boloursaz Mashhadi,~\IEEEmembership{Senior Member, IEEE}, \\ and Rahim Tafazolli,~\IEEEmembership{Fellow, IEEE}
\thanks{
F.~Zeinali, M.~Boloursaz Mashhadi, and R.~Tafazolli are with 5GIC $\&$ 6GIC,~Institute for Communication Systems (ICS),~University of Surrey,~UK (e-mail: \{f.zeinali, m.boloursazmashhadi, r.tafazolli\}@surrey.ac.uk).

}}

\markboth{}%
{Shell \MakeLowercase{\textit{et al.}}: Bare Demo of IEEEtran.cls for IEEE Journals}

\maketitle

\begin{abstract}
\textit{Token Communications (TokenCom)} has recently emerged as an effective new paradigm, where \textit{tokens are the unified units of multimodal communications and computations}, enabling efficient digital semantic and goal-oriented communications in future wireless networks. To establish a shared semantic latent space, the transmitters/receivers in TokenCom need to agree on an identical tokenizer model and codebook. To this end, an initial \textit{Tokenizer Agreement (TA)} process is carried out in each communication episode, where the transmitter/receiver cooperate to choose from a set of pre-trained tokenizer models/codebooks available to them both for efficient TokenCom. In this letter, we investigate TA in a multi-user downlink wireless TokenCom scenario, where the base station equipped with multiple antennas transmits video token streams to multiple users. We formulate the corresponding mixed-integer non-convex problem, and propose a hybrid reinforcement learning (RL) framework that integrates a deep Q-network (DQN) for joint tokenizer agreement and sub-channel assignment, with a deep deterministic policy gradient (DDPG) for beamforming. Simulation results show that the proposed framework outperforms baseline methods in terms of semantic quality and resource efficiency, while reducing the freezing events in video transmission by 68\% compared to the conventional H.265-based scheme.

%Semantic communication (SemCom) has emerged as a promising paradigm to enhance communication efficiency by transmitting the meaning of information rather than raw bits.

%Efficient TA in multi-user wireless TokenCom involves mixed continuous and discrete decision variables, including choice of the tokenizer, to be jointly optimized with beamforming, power allocation, and sub-channel assignment.

%maximize a utility function that jointly captures semantic quality, spectral efficiency, and power consumption. 
\end{abstract}


\IEEEpeerreviewmaketitle
\vspace{-1.05em}

\section{Introduction}
The recent integration of Large AI Models (LAMs) and Multimodal Large Language Models (MLLMs) with wireless networks provides ample opportunities to develop innovative technologies with transformative potential. One such technology is \textit{Token Communications (TokenCom)} \cite{11175596, qiao2025todma}, that leverages the scaling and generalization capabilities of LAMs/MLLMs, to develop bandwidth efficient ultra-low-bitrate semantic- and goal-oriented communications. Tokens are the basic processing units of text, images, audio, and video signals in state-of-the-art LAMs/MLLMs, and TokenCom adopts tokens as the universal semantic carrying units that  generalize beyond tasks, datasets, and signal modalities, aiming to address the \textit{lack of generalizability} gap in the conventional semantic- and goal-oriented communications \cite{9955312, 9955525, 10872776}. In TokenCom, multimodal signals are first tokenized to a stream of tokens, and then transmitted via their indices in a pre-trained \textit{tokenizer codebook}, shared between the transmitter/receiver. Thereby, to establish a shared semantic latent space, the transmitter and receiver carry out an initial \textit{Tokenizer Agreement (TA)} process in each communication episode to choose from a set of standardized pre-trained tokenizer models/codebooks assumed available to both sides. Subsequently, TokenCom can flexibly adapt to varying channel/network conditions or generalize to new tasks, by a simple change of the tokenizer/de-tokenizer model/codebook pair at the transmitter/receiver, making efficient TA crucial for TokenCom.

% Moreover, adaptive TA enables achieving various rate-distortion-perception performance [?]-[?], thereby making TokenCom flexibly adaptable under varying channel and network conditions.

In a multi-user wireless TokenCom setup, the choice of the tokenizer should be jointly optimized with sub-channel assignment, beamforming, and resource allocation, which leads to a mixed-integer non-convex problem. The conventional optimization techniques would incur high computational complexity for this problem due to the need for numerous iterations of convex relaxations, but obtain only a locally optimal solution. Moreover, dynamic changes in the multi-user wireless channels, or changes in the users' semantic requirements, lead to performance loss when using the conventional optimizations. This makes Reinforcement Learning (RL) most suitable for TA in multiuser wireless TokenCom, enabling adaptability to channel conditions and semantic requirements.

%to the performance of TokenCom framework. balance semantic quality, spectral efficiency, and energy consumption in practical SemCom systems. Moreover, by changing the tokenizer model/codebook, the transmitter/receiver can achieve various points on the rate-distortion-perception curve [?], thereby adapting the datarate to varying channel/network conditions. TA introduces new opportunities and challenges. 

%Specifically, with a change of the tokenizer, we affects the end-to-end semantic quality as different tokenizer achieve various points on the rate-distortion-perception curve.  

%Semantic communication (SemCom) has recently emerged as a promising paradigm to address the spectrum scarcity and efficiency limitations of traditional wireless communication systems~\cite{10872776}. By focusing on transmitting meaning rather than raw bits, SemCom enables significant bandwidth savings, enhances robustness under adverse channel conditions, and improves overall communication efficiency. However, these advantages introduce new challenges in resource allocation, as semantic encoding and decoding models have dynamic requirements for power, bandwidth, and computational resources depending on the task and user type. Efficient resource allocation is therefore crucial to balance semantic quality, spectral efficiency, and energy consumption in practical SemCom systems.  

Several recent studies have proposed resource allocation and adaptation schemes for wireless semantic communication (SemCom) systems. In~\cite{10122232}, RL was used to dynamically allocate resources in a task-oriented SemCom network, prioritizing high-value semantic data to maximize long-term task transmission efficiency. In~\cite{10981779}, a proximal policy optimization (PPO)-based framework was proposed for wireless semantic image transmission, optimizing semantic spectral efficiency while maintaining acceptable image reconstruction quality. Textual semantic communication was explored in~\cite{9832831}, where attention-enhanced PPO algorithms jointly allocate resource blocks and select key semantic triples to maximize semantic similarity. For coexisting semantic and bit-level communications,~\cite{10845882} optimized beamforming to maximize the semantic rate while meeting the quality of service (QoS) requirements. % Finally,~\cite{9763856} focused on text SemCom, leveraging a deterministic optimization approach for semantic spectral efficiency and semantic similarity.} 

These existing studies have focused on the conventional SemCom schemes and typically use smaller AI models without token-based signal processing. The need to migrate to LAMs/MLLMs and the additional adaptive TA process required in TokenCom, introduce new challenges, which we study in this work. Specifically, the choice of the pre-trained tokenizer model/codebook introduces coexistence of discrete and continuous decision variables, which makes the optimization problem highly complex and non-convex, limiting the effectiveness of conventional RL techniques. Moreover, these prior studies have mostly considered single-user image or text SemCom, while we consider multiuser video TokenCom leveraging a subset of state-of-the-art pre-trained video tokenizers. Table \ref{tab:related_works} provides an overview of the key differences between this work and the literature on RL-assisted SemCom. To the best of our knowledge, this work is the first of its kind to tackle the multi-user wireless TokenCom problem, proposing a new hybrid DQN-DDPG RL framework for joint adaptive tokenizer agreement, with sub-channel assignment, beamforming, and resource allocation. The proposed hybrid DQN-DDPG solution integrates a DQN agent for discrete TA and sub-channel assignment, with a DDPG agent for continuous beamforming, allowing efficient learning in a mixed-action space. Simulation results demonstrate significant improvements in semantic quality outperforming benchmarks, while enabling adaptive, resource-efficient multiuser wireless TokenCom.

\begin{table}[t]
\centering
\caption{Summary of related works in wireless SemCom.}
\label{tab:related_works}

\begingroup
\scriptsize
\renewcommand{\arraystretch}{0.9}
\setlength{\tabcolsep}{2pt}

\begin{tabular}{|
m{0.12\columnwidth}<{\centering}|
m{0.28\columnwidth}<{\centering}|
m{0.34\columnwidth}<{\centering}|
m{0.22\columnwidth}<{\centering}|
}
\hline
\textbf{} & \textbf{Objective} & \textbf{Resource Adaptation} & \textbf{Solution} \\
\hline
\cite{10122232} & Task Efficiency & Power/Subchannel/Compression  & DDPG \\
\hline
\cite{10981779} & Semantic Spectral Efficiency & Power/Subchannel/Compression & PPO with clipping \\
\hline
\cite{9832831} & Semantic Similarity & Subchannel/Semantic Triple Selection & PPO with attention \\
\hline
\cite{10845882} & Semantic Transmission Rate & Beamforming/Compression & MM-FP and LP-MM-FP \\
\hline
This Work & Token Resource--Quality Trade-off & Subchannel/Beamforming/ Tokenizer Agreement (TA) & Hybrid DQN-DDPG \\
\hline
\end{tabular}

\endgroup

\end{table}
\vspace{-0.25cm}

\section{Multi-User Wireless Token Communications}
%\footnote{The proposed multi-user wireless TokenCom framework is extendable to other modalities, e.g., image, audio, etc.}

\subsection{Tokenizer Agreement for Adaptive Video TokenCom}
Assume a piece of video $\boldsymbol{V} \in \mathbb{R}^{F \times H \times W \times C}$, where $F$ is the number of temporal frames, $H$ and $W$ represent height and width in pixels, and $C=3$ is the number of  RGB channels. A general tokenizer (encoder) compressing the video with a compression factor $\boldsymbol{\mu} =  (\mu_F,\mu_H,\mu_W) = (\frac{F^{\prime}}{F}, \frac{H^{\prime}}{H}, \frac{W^{\prime}}{W})$, produces latent representation $\boldsymbol{z} = \zeta_{\text{enc}}(\boldsymbol{V}) \in \mathbb{R}^{F^{\prime} \times H^{\prime} \times W^{\prime} \times C^{\prime}}$, which contains $F^{\prime} \times H^{\prime} \times W^{\prime}$ token embedding vectors of size $C^{\prime} \times 1$. Token embeddings are then mapped to integer token IDs that can be processed by transformers for video modeling, prediction, or generation. Each token ID represents the token index from a pre-trained token codebook.
%Depending on the tokenizer quantization scheme, each token ID may represent the token index from a pretrained token codebook, i.e. in Vector Quantization (VQ)-based tokenization, or a quantized scalar tuple from a learned multidimensional quantizer representing a finite token vocabulary, i.e. in Finite Scalar Quantization (FSQ)-based tokenization. In the case of FSQ, each scalar in the latent embedding vector is quantized independently into a finite number of bins, the combination of which forms the unique ID from the token vocabulary, without requiring to store a large token codebook. 
The resulting stream of token IDs then serves as an ultra-low-bitrate, super compact semantic representation of the video to be transmitted over the wireless channel. The receiver can then reconstruct the video from the received token IDs using the corresponding de-tokenizer (decoder) $\hat{\boldsymbol{V}}=\zeta_{\mathrm{dec}}(\hat{z})$. Note that state-of-the-art tokenizer/de-tokenizer pairs are pre-trained jointly on large corpora of data, thereby each tokenizer is compatible only with its corresponding de-tokenizer, and the pair should be applied together, requiring a tokenizer agreement process between the transmitter and receiver to establish TokenCom.

The choice of tokenizer/de-tokenizer pair determines the size of the latent representation and the token codebook/vocabulary, hence the resulting compression rate in bits per pixel (bpp) and the wireless resources required for TokenCom. The compression rate is calculated as $\eta=\mu_F \times \mu_H \times \mu_W \times \log_2|O|$, where $|O|$ denotes the size of the token codebook/vocabulary. Accordingly, different tokenizers achieve different rate-distortion/perception performance, i.e., $q=f(\eta)$, where $q$ is a non-increasing distortion/perception function of the tokenizer compression rate $\eta$ \cite{RDP1, RDP2} defined in terms of any reconstruction/synthesis distortion/perception metrics, e.g., PSNR, SSIM, rFVD, LPIPS, etc. The rate required in bits per second (bps) is $\rho \times \eta \times H \times W$, in which $\rho$ is the number of frames per second (fps) for the video.

To enable rate-distortion/perception adaptive TokenCom in a multiuser wireless downlink setup, we consider that a tokenizer agreement process is carried out between the base station (BS) and all users upon initiation of each communication episode. We assume that the BS has access to a comprehensive set of pre-trained tokenizers $\mathcal{T}$, while due to hardware constraints, each user has local access to a relatively more limited set of pre-trained de-tokenizers $\mathcal{D}_i$, pre-installed or cached on demand. For downlink TA in each communication episode, the following two step process is carried out

% , during the connection initiation process
\textbf{-Step 1:} Each user $i \in \{1, ..., U\}$ informs the BS of its available set of de-tokenizers via their name tags, e.g. \{``Cosmos-0.1-Tokenizer-DV8$\times$16$\times$16", ``LlamaGen-Tokenizer
 8x8"\}, in a short message. The BS then generates $\mathcal{M}_i$, which is the set of all compatible tokenizer/de-tokenizer pairs for TokenCom with each user $i$, as follows:
\[
\mathcal{M}_i \! = \! \big\{ \big( \zeta_{\mathrm{enc}}^{(m)},\, \zeta_{\mathrm{dec}}^{(m)} \big) | \zeta_{\mathrm{enc}}^{(m)} \! \in \! \mathcal{T}, \zeta_{\mathrm{dec}}^{(m)} \! \in \! \mathcal{D}_i \big\}, m \! \in \! \{1,\dots,M_i\}.
\]

\begin{figure}[t]
\centering
    \includegraphics[width=0.93\linewidth]{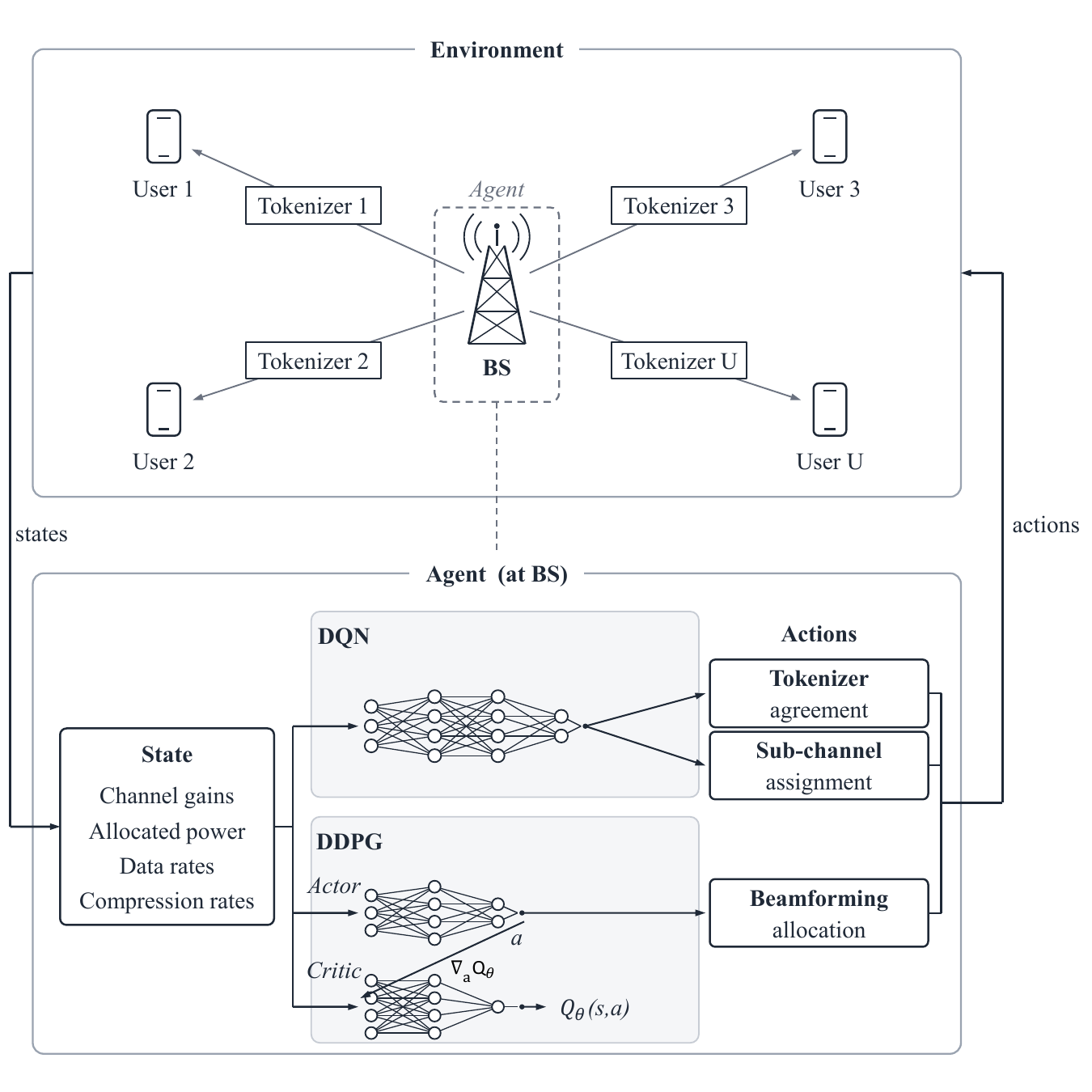}
\caption{Multi-User Wireless TokenCom with Adaptive RL-based Tokenizer Agreement}
\label{systemmodel}
\end{figure}

\textbf{-Step 2:} The BS adaptively optimizes the best tokenizer/de-tokenizer pair index $m_i \in \{1,\dots,M_i\}$, based on the channel conditions and available resources in each communication episode, and communicates the de-tokenizer name tag corresponding to the selected pair $m_i$ back to each user. After this, the TA process is concluded, and a shared semantic latent embedding space with compression rate $\eta_{m_i}$ corresponding to the $m_i$th pair is established between the BS and each user.

%\[
%\boldsymbol{z}_i = \zeta_{\text{enc}}^{m}(I_i),
%\] 

%This abstraction reflects the agreement that both BS and receivers know how to encode and decode according to the chosen model. The latent vector $\boldsymbol{z}_i$ is partitioned into multiple parts and mapped onto the RBs allocated to user $i$. 

\subsection{Joint Tokenizer Agreement, Resource Allocation, and Beamforming for TokenCom}
We consider a multi-user downlink wireless TokenCom system in which a BS equipped with $N$ transmit antennas serves $U$ single-antenna users over $R$ orthogonal resource blocks (RBs), each with bandwidth $B$. Each communication episode consists of $T$ time slots. We define a binary allocation indicator variable $\kappa_{i}^t[l]$, which is equal to 1 if RB $l$ is assigned to user $i$ in time slot $t$, and is 0 otherwise. For each user $i$, the set of RBs allocated to it at time slot $t$ is $\mathcal{R}^t_i = \{ l | \kappa_i^t[l] = 1\}$ where $|\mathcal{R}^t_i|=\sum_{l=1}^{R} \kappa_i^t[l]$ is the number of RBs assigned to user $i$ at time slot $t$. In each time slot each RB can be simultaneously assigned to at most $\kappa$ users to limit the interference, i.e., $\sum_{i=1}^{U} \kappa_i^t[l] \;\leq\; \kappa, \forall l\in\{1,\dots,R\}$. On each allocated RB $l$, the BS applies a beamforming vector $\boldsymbol{w}^t_i[l] \in \mathbb{C}^{N}$ at each time slot $t$, thereby the signal-to-interference-plus-noise ratio (SINR) for user $i$ in RB $l$ is
\begin{equation}
\mathrm{SINR}^t_i[l] = \kappa^t_i [l] \frac{|\boldsymbol{h}_i^t[l]^{H}\,\boldsymbol{w}^t_i[l]|^{2}}{\Sigma_{j\neq i} \kappa^t_j [l] |\boldsymbol{h}_i^t[l]^{H}\,\boldsymbol{w}^t_j[l]|^{2}+B\,N_0}.
\end{equation} 
The achievable rate for user $i$ on RB $l$ at time slot $t$ is then given by $R_i^t[l] = B \log_2\!\big(1+\mathrm{SINR}_i^t[l]\big)$, and the total achievable rate for user $i$ at time slot $t$ is the sum over all its allocated RBs $R_i^t = \sum_{l \in\mathcal{R}_i^t} R_i^t[l]$. The transmit power allocated to user $i$ on RB $l$ at time slot $t$ is given by $p_i^t[l] = \|\boldsymbol{w}_i^t[l]\|_2^2$, thereby summing over all RBs, the power allocated to user $i$ at time slot $t$ is given by $P_i^t = \sum_{l\in\mathcal{R}_i^t} \|\boldsymbol{w}^t_i[l]\|_2^2$.

Efficient beamforming and optimum allocation of RBs increase the communication rate $R_i^t$, enabling to adopt a tokenizer/de-tokenizer pair with improved distortion/perception quality for video TokenCom for each user. The distortion/perception quality achieved for user $i$ is then given by $q_i^t=f(\eta_{m_i})$. We normalize $q_i^t$ using $\bar{q}_i^t=\frac{q_i^t-q_{\text{min}}}{q_{\text{max}}-q_{\text{min}}}$ for better numerical stability, where a larger $\bar{q}_i^t$ represents an improved distortion/perception quality. Finally, for a controllable balance to maximize the distortion/perception quality while saving on the sum power allocated to users, we define the system utility in time slot $t$ as $\mathcal{U}^t = \frac{1}{U}\sum_{i=1}^{U} \left( 
\alpha \bar{q}_i^t - \beta \frac{P_i^t}{P_{\text{BS}}}
\right)$, where $\alpha, \beta > 0$ are tunable weights. We assume sufficient computational capabilities available at the BS and users' device hardware to accommodate any of the tokenizer/de-tokenizer pairs in $\mathcal{M}_i$.  

The joint optimization of tokenizer agreement, RB allocation, and beamforming aims to maximize the system utility while satisfying the resource constraints is given as follows:
\begin{subequations}\label{eq:opt}
\begin{align}
&\max_{\{m_i\},\,\{\kappa_i^t[l]\},\,\{\boldsymbol{w}^t_i[l]\}}
    \; \sum_{t=1}^{T}\mathcal{U}^t \\[1mm]
\text{s.t.}\;\;
& \sum_{i=1}^{U}\!\sum_{l=1}^{R}\!\|\boldsymbol{w}_i^t[l]\|_2^2 \le P_{\mathrm{BS}},\;\forall t,
&& \!\!\!\!\!\!\!\!\!\!\!\!\!\!\!\!\!\! \sum_{i=1}^{U}\kappa_i^t[l] \le \kappa,\;\forall l,t, \label{eq:opt_b,opt_a}
\\[0.5mm]
& R_i^t \ge \rho \eta_{m_i}HW,\;\forall i,t,
&& \!\!\!\!\!\!\!\!\!\!\!\!\!\!\!\!\!\! q_i^t \ge q_{\min},\;\forall i,t, \label{eq:opt_g,opt_e}
\\[0.5mm]
& K_{\min} \le \sum_{l=1}^{R}\kappa_i^t[l] \le K_{\max},\;\forall i,t,
&& R_i^t \ge R_{\min},\;\forall i,t, \label{eq:opt_f,opt_h}
\\[0.5mm]
& m_i \in \{1,\dots,M_i\},\;\forall i,
&& \!\!\!\!\!\!\!\!\!\!\!\!\!\!\!\!\!\!\!\!\!\! \kappa_i^t[l] \in \{0,1\},\;\forall i,l,t, \label{eq:opt_c,opt_d}
\end{align}
\end{subequations}
where, the two constraints in~\eqref{eq:opt_b,opt_a} reflect the total transmit power budget of the BS across all users and RBs, and ensure that each RB is assigned to at most $\kappa$ users to limit the interference in each time slot. The two constraints in~\eqref{eq:opt_g,opt_e} ensure that the Shannon rate allocated to each user $i$ can support transmission of the video content for that user, while the video transmission quality for every user remains above the minimum acceptable level. The two constraints in~\eqref{eq:opt_f,opt_h} ensure that the number of RBs allocated to each user vary within an acceptable range, and enforce the minimum data rate requirement for the users. Finally, the two constraints in~\eqref{eq:opt_c,opt_d} represent the tokenizer agreement requirement, enforcing that the BS and users agree on one of the predefined compatible tokenizer/de-tokenizer pairs, and reflect the binary nature of sub-channel assignment variables. The above is a mixed-integer non-convex problem which requires development of advanced optimization algorithms to find a near-optimal solution for complex time varying wireless environments in an adaptive computationally efficient manner. In the next section, we provide our proposed hybrid DQN-DDPG algorithm for this optimization.
\vspace{-0.4cm}
\section{Proposed Hybrid DQN--DDPG Algorithm}
In this section, we present our proposed RL algorithm for joint tokenizer agreement, resource allocation and beamforming. We formulate the problem as a Markov decision process (MDP) defined by the tuple $(\mathcal{S}, \mathcal{A}, P, r, \gamma)$, where $\mathcal{S}$ is the state space, $\mathcal{A}$ is the action space, $P(s'|s,a)$ is the transition probability, $r(s,a)$ is the reward function, and $\gamma \in [0,1)$ is the discount factor. The tokenizer/de-tokenizer pair is selected once at the beginning of each episode for all users, while sub-channel allocation/beamforming are updated every time slot.
\vspace{-0.8cm}
\subsection{Action Space}
The action space is structured as $a^t = \Big(\{m_i\}_{i},\ \{\kappa_i^t[l]\}_{i,l},\ \{\boldsymbol{w}_i^t[l]\}_{i,l}\Big),$ where, $m_i \in \{1,\dots,M_i\}$ is the tokenizer model index chosen at the start of each episode, and is fixed during all time slots. For every time slot, the DQN branch outputs RB assignment indicators $\kappa_i^t[l]$, while the DDPG branch generates the beamforming vectors $\{\mathbf{w}_i^{\,t}[l]\}$ for
    all users and RBs. As each $\mathbf{w}_i^{\,t}[l]\in\mathbb{C}^{N}$ is
    represented by its real and imaginary parts, the actor output dimension
    is $U\!\times\!R\!\times\!2N$, which scales with both the number of
    antennas $N$ and users $U$; accordingly, the actor/critic are
    instantiated with the matching dimensions for each $(N,U)$ setting. The
    $2N$ real values produced per $(i,l)$ pair are split into two halves that
    form the real and imaginary parts of $\mathbf{w}_i^{\,t}[l]$.
\vspace{-0.4cm}
\subsection{State Representation}
The state $s^t$ observed at each slot includes user channels, transmission powers, data, and compressions rates as 
$s^t = \Big\{ \{\boldsymbol{h}_i^t[l]\}_{i,l},\ \{p_i^t[l]\}_{i,l},\ \{R_i^t\}_i,\,\ \{\eta_{m_i}\}_i \Big\}.$
The inclusion of $\eta_{m_i}$ in the state ensures that the RL agent conditions its per-step decisions on the chosen tokenizer/de-tokenizer pairs for the current episode.
\vspace{-0.4cm}
\subsection{Reward Function}
The instantaneous reward follows the utility function $r^t = \mathcal{U}^t - \sum_{n=1}^{8} \lambda_{\text{pen}} \Upsilon_n^t,$ where $\Upsilon_n^t$ = 1, $n \in \{1,2,...,8\}$ if constraint $n$ in time slot $t$ is not satisfied, and 0 otherwise. Since $\{m_i\}_i$ is fixed during an episode, the impact of TA on semantic quality is reflected in the per-step reward through $q^t_i$.%$\widehat{\mathrm{PSNR}}_i = f(R_i,\theta_{i}^{m})$.
\begin{algorithm}[t]
\caption{Hybrid DQN--DDPG for Joint Tokenizer Selection and Resource Allocation}
\begin{algorithmic}[1]
\State \textbf{Initialize:} Set Environment $\mathcal{E}$ and learning parameters.

\For{$e=1$ \textbf{to} $E$}
    \State Reset environment $\mathcal{E}$, obtain initial state $s^1$.
    \State \textbf{Tokenizer selection at episode start:}
        \Statex \hspace{0.6cm} With prob. $\epsilon$: select random tokenizer $m_i$;
        \Statex \hspace{0.6cm} otherwise: $m_i \leftarrow \arg\max Q_\phi(s^1)$.
        \Statex \hspace{0.6cm} Set tokenizer/de-tokenizer pair $m_i$ in $\mathcal{E}$.
    \For{$t=1$ \textbf{to} $T$}
        \State Observe $s^t$.
        \State RB allocation by DQN,
            With prob. $\epsilon$: choose 
            
            random $\kappa^t$;
            otherwise: $\kappa^t \leftarrow \arg\max Q_\phi(s^t)$.
        \State Beamforming by DDPG,
            $\boldsymbol{w}^t \leftarrow \pi_\psi(\bar{s}^t) + \mathcal{N}^t$, 
            
            with exploration noise $\mathcal{N}^t$.
        \State Execute $(\kappa^t, \boldsymbol{w}^t)$ in $\mathcal{E}$;
            observe $s^{t+1}$.
        \State Compute reward and Store $(s^t, (m_i,\kappa^t), r^t, s^{t+1})$ in $\mathcal{B}_1$ and Store $(s^t, \boldsymbol{w}^t, r^t, s^{t+1})$ in $\mathcal{B}_2$.
        \If{$|\mathcal{B}_1| \geq \text{batch\_size}$ and $|\mathcal{B}_2| \geq \text{batch\_size}$}
            \State Sample mini-batch from $\mathcal{B}_1$ and $\mathcal{B}_2$.
            \State Update DQN params $\phi$ with TD loss.
            \State Update Critic $\theta$ by~\eqref{CRITICloss}, and actor $\psi$ by $L_{\text{actor}} = - Q_\theta(s^t, \pi_\psi(s^t))$.
            \State Soft update targets by Polyak averaging. %(\ref{target_critic}).
        \EndIf
        \State $s^t \leftarrow s^{t+1}$.
    \EndFor
    \State $\epsilon \leftarrow \max(\epsilon_{\text{end}}, \epsilon \cdot \epsilon_{\text{decay}})$.
\EndFor
\end{algorithmic}
\end{algorithm}
\vspace{-0.8cm}
\subsection{Learning Framework}
As shown in Fig.\ref{systemmodel}, at each time slot $t$, the RL agent observes state $s^t \in \mathcal{S}$, selects an action $a^t \in \mathcal{A}$, receives an immediate reward $r^t$, and the environment transitions to $s^{t+1}$. The objective is to maximize the expected discounted return $J = \mathbb{E}\left[\sum_{t=1}^{\infty} \gamma^t r^t \right]$.
%\begin{equation}
%    J = \mathbb{E}\left[\sum_{t=1}^{\infty} \gamma^t r^t \right].
%\end{equation}

The discrete actions consist of tokenizer selection once at the beginning of each episode, and RB allocation at each slot. The DQN approximates the action-value function
\begin{equation}
    Q^{\pi}(s,a) = \mathbb{E}\left[ r^t + \gamma \max_{a'} Q_\phi(s^{t+1},a') \,\big|\, s^t=s, a^t=a \right].
\end{equation}
The DQN is trained minimizing the temporal-difference (TD) 
$\mathcal{L}_{\text{DQN}}(\phi) = \mathbb{E}_{(s,a,r,s') \sim \mathcal{B}} \Big[ \big(Q_\phi(s,a) - \upsilon^t \big)^2 \Big],$
where $\phi$ denotes the parameters of the neural network. The target value is defined as $\upsilon^t = r + \gamma \max_{a'} Q_{\phi^{\prime}}(s',a')$, where $\phi^{\prime}$ denotes the parameters of the target network. %updated periodically.
%\begin{equation}
%    \upsilon^t = r + \gamma \max_{a'} Q_{\phi^{\prime}}(s',a'),
%\end{equation}

%\begin{equation} 
%    \upsilon^t = r + \gamma Q_{\omega^{\prime}}(s', \pi_{\psi^\prime}(s')),
%\end{equation}

The continuous beamforming vectors are generated using an actor-critic architecture. The actor $\pi_\psi(s)$ with parameters $\psi$ maps states to continuous actions, while the critic $Q_\theta(s,a)$ with parameters $\theta$ evaluates their quality. The critic is trained by minimizing the Bellman loss
\begin{equation} \label{CRITICloss}
    \mathcal{L}_{\text{Critic}}(\theta) = \mathbb{E}_{(s,a,r,s') \sim \mathcal{B}} \Big[ \big(Q_\theta(s,a) - \upsilon^t \big)^2 \Big],
\end{equation}
with target $\upsilon^t = r + \gamma Q_{\theta^{\prime}}(s', \pi_{\psi^\prime}(s'))$, where $\theta^{\prime}$ and $\psi^{\prime}$ are the target critic and actor parameters. The actor is updated by applying the deterministic policy gradient $\nabla_\psi J(\psi) = \mathbb{E}_{s \sim \mathcal{B}} \Big[ \nabla_a Q_\theta(s,a)\big|_{a=\pi_\psi(s)} \, \nabla_\psi \pi_\psi(s) \Big].$
\begin{comment}
To stabilize training, target networks are updated using the Polyak averaging scheme $ \phi^{\prime} \leftarrow \tau \phi + (1-\tau)\phi^{\prime}$, $\psi^{\prime} \leftarrow \tau \psi + (1-\tau)\psi^{\prime}$, and $\theta^{\prime} \leftarrow \tau \theta + (1-\tau)\theta^{\prime}$, 
%\begin{align}
%    \phi^{\prime} \leftarrow \tau \phi + (1-\tau)\phi^{\prime},
%    \psi^{\prime} \leftarrow \tau \psi + (1-\tau)\psi^{\prime}, 
%    \omega^{\prime} \leftarrow \tau \omega + (1-\tau)\omega^{\prime},\label{target_critic}
%\end{align}
where $\tau \ll 1$ is the soft update coefficient. All transitions $(s^t,a^t,r^t,s^{t+1})$ are stored in a replay buffer $\mathcal{B}$ of finite capacity. At each update, a mini-batch is uniformly sampled from $\mathcal{B}$ to de-correlate samples and improve learning stability. 
\end{comment}
The proposed hybrid DQN-DDPG is summarized in \textbf{Algorithm 1}. The complexity of the proposed
framework is dominated by the forward/backward passes of the DQN,
actor, and critic networks of the DDPG. Since the state is dominated by the per-RB
channel vectors $\{h_i^t[l]\}$, its dimension is $\mathcal{O}(URN)$;
likewise, the actor output (beamforming vectors $\{w_i^t[l]\}$) is of
dimension $\mathcal{O}(URN)$, while the factored DQN output for RB
assignment and tokenizer selection is $\mathcal{O}(UR+\sum_i M_i)$. For
MLPs with $L$ hidden layers of width $D$, a single forward pass of any
of the three networks costs $\mathcal{O}(URND+LD^2)$.
\vspace{-0.4cm}
\section{Simulation Results}

We consider a multi-user wireless video TokenCom setup over Rayleigh block fading channels, with a subcarrier spacing of $B=30~\text{kHz}$, and parameter setting as provided in Table~\ref{settings}. We assume that the BS and all users can support a set of $M_i=4$ state-of-the-art pre-trained discrete video tokenizer/de-tokenizer pairs with different compression rates and semantic distortion/perception quality levels, as summarized in Table~\ref{tokenizers}. All results are reported on the DAVIS video dataset with $\rho\!=\!24$ fps. We compare our proposed wireless TokenCom framework with DQN-DDPG based adaptive TA, with the following 4 baselines 
\begin{enumerate}[]
		\item \textbf{DDPG-TA:} TokenCom using a conventional DDPG algorithm with discretized outputs for adaptive TA.
        \item \textbf{Agnostic-TA:} TokenCom using the same tokenizer for all users, i.e., TA agnostic to heterogeneity in the users' channel conditions.
        \item \textbf{Fixed-TA:} TokenCom with a non-adaptive TA, i.e., the tokenizer is fixed in all the communication episodes.
        \item \textbf{Conventional:} This is a conventional communication baseline using the digital H.265 video codec \cite{6317156} with rate adaptation, as well as DQN-DDPG for joint resource allocation and beamforming.
	\end{enumerate}

%\begin{figure}[t]
    %\centering
    %\begin{subfigure}[t]{0.8\linewidth}
    %    \centering
    %    \includegraphics[width=\linewidth]{eps/reward.eps}
    %    \caption{\footnotesize Reward performance.}
    %    \label{reward}
    %\end{subfigure}
    %\hfill
    %\begin{subfigure}[t]{0.8\linewidth}
    %    \centering
    %    \includegraphics[width=\linewidth]{eps/freezing.eps}
        %\caption{\footnotesize Freezing performance.}
        %\label{freezing}
    %\end{subfigure}
    %\caption{\footnotesize Convergences and freezing performance.}
    %\label{converge}
%\end{figure}

\begin{comment}
\begin{figure}[t]
	\includegraphics[width=0.75\linewidth]{eps/freezing.eps}
	\centering
	\caption{\footnotesize Convergences and freezing performance.}
    \label{converge}
\end{figure}

\begin{figure}[t]
	\includegraphics[width=0.75\linewidth]{eps/freezing_quality.eps}
	\centering
	\caption{\footnotesize Average freezing over qualities.}
	\label{freezing_quality}
\end{figure}

\begin{figure}[t]
	\includegraphics[width=0.75\linewidth]{eps/PSNR_user.eps}
	\centering
	\caption{\footnotesize Average PSNR versus number of users.}
	\label{PSNR_users}
\end{figure}

\begin{figure}[t]
	\includegraphics[width=0.75\linewidth]{eps/PSNR_power.eps}
	\centering
	\caption{\footnotesize Average PSNR versus transmit power.}
	\label{PSNR_power}
\end{figure}
\end{comment}

In this simulation, we train the RL agent using parameter settings summarized in Table~\ref{settings}. For performance evaluation, we use the video freezing rate (\%) and a normalized quality score that combines PSNR and SSIM, each min--max normalized to $[0,1]$ and averaged; any other distortion/perception metric expressed as a function of the compression rate can be used similarly. Freezing of the video can occasionally occur for a user in a time step, if the wireless transmission data rate for that user falls below the bitrate required for its selected tokenizer, i.e., ~\eqref{eq:opt_g,opt_e} is unsatisfied in that time step.

%In Fig.\ref{reward}, %In contrast, the Same-TA configurations exhibit lower and more fluctuating rewards.

%In terms of convergence, our proposed hybrid RL framework with adaptive TA achieves the highest and most stable reward convergence, outperforming baselines. Incorporating adaptive TA enables the hybrid DQN–DDPG agent to learn consistent semantic–resource mappings, resulting in stability and higher long-term rewards. 

\begin{figure}[t]
    \centering
    \begin{subfigure}[t]{0.48\linewidth}
        \centering
        \includegraphics[width=\linewidth]{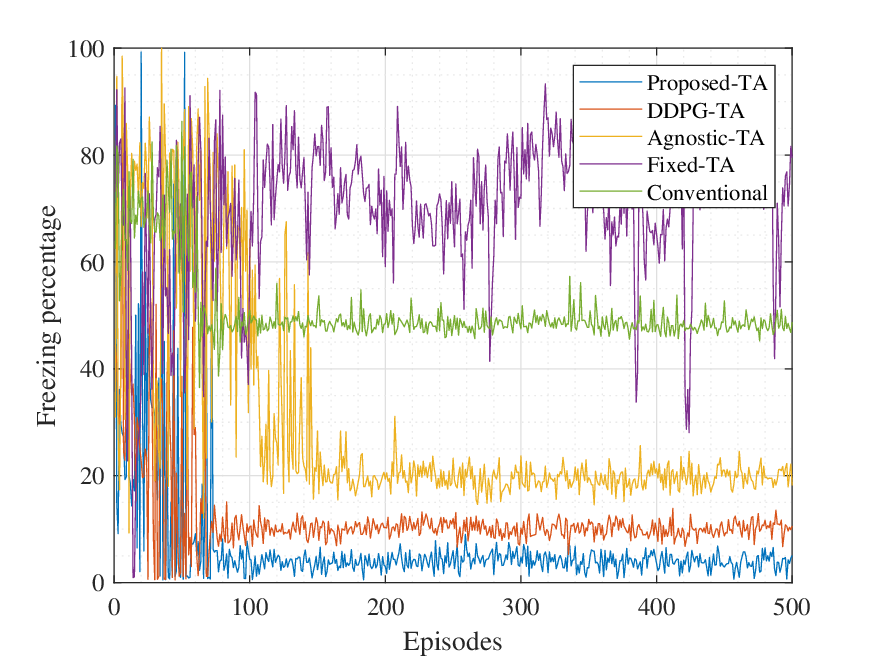}
        \caption{\footnotesize Freezing vs. episodes.}
        \label{fig:converge}
    \end{subfigure}
    \hfill
    \begin{subfigure}[t]{0.48\linewidth}
        \centering
        \includegraphics[width=\linewidth]{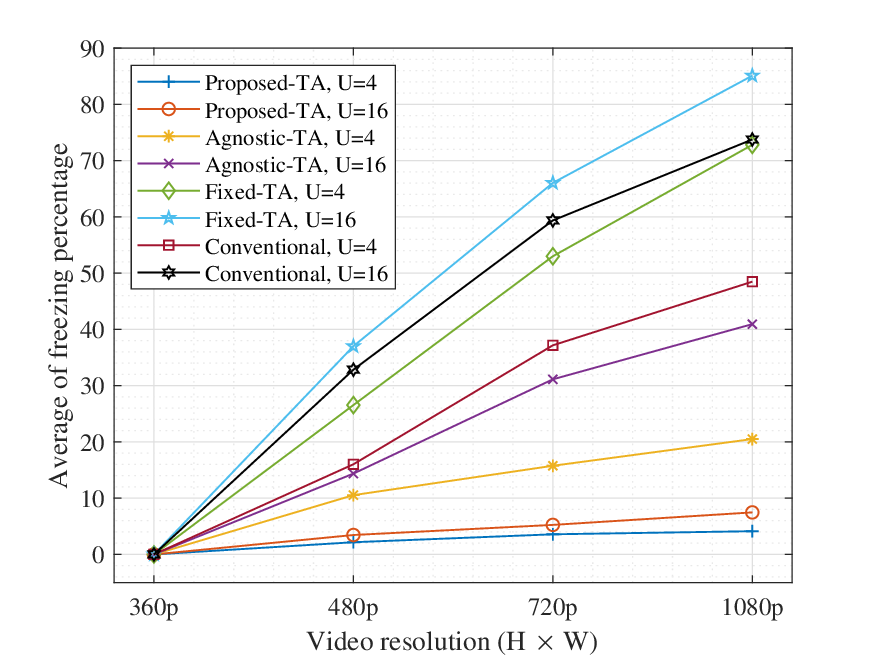}
        \caption{\footnotesize Freezing vs. video resolution.}
        \label{fig:freezing_quality}
    \end{subfigure}

    \begin{subfigure}[t]{0.48\linewidth}
        \centering
        \includegraphics[width=\linewidth]{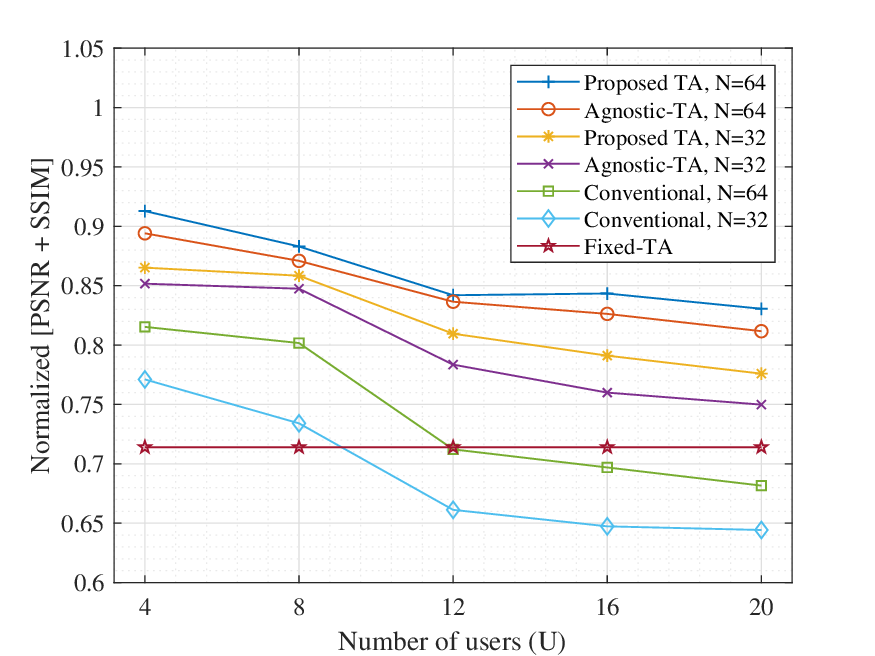}
        \caption{\footnotesize Semantic quality vs. $U$.}
        \label{fig:PSSI_users}
    \end{subfigure}
    \hfill
    \begin{subfigure}[t]{0.48\linewidth}
        \centering
        \includegraphics[width=\linewidth]{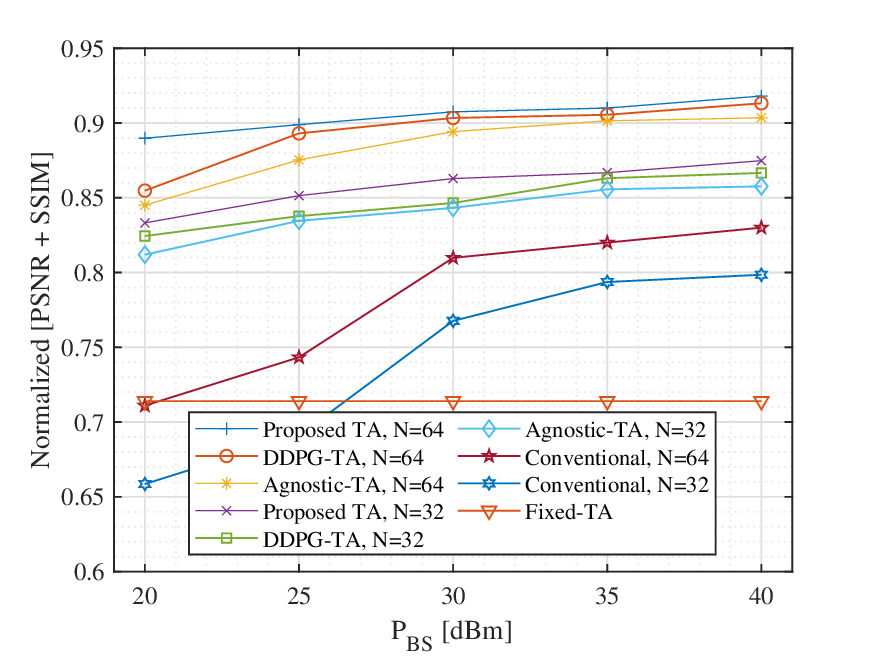}
        \caption{\footnotesize Semantic quality vs.  power.}
        \label{fig:PSSI_power}
    \end{subfigure}

    \caption{\footnotesize Performance results.}
    \label{fig:all_subfigs}
\end{figure}

Fig.\ref{fig:converge} depicts the freezing rate reduction versus training episodes for different baselines. The proposed TokenCom framework achieves the lowest and most stable freezing rate reduction after a brief initial exploration phase, indicating that our proposed DQN-DDPG agent effectively learns to adaptively match tokenizers to users’ demands and channel conditions, specifically improving over pure DDPG-TA. The Agnostic-TA and Fixed-TA baselines exhibit higher freezing levels due to limited adaptability to heterogeneous user channel conditions, specifically, Fixed-TA suffers from persistently high and oscillatory freezing. %due to its lack of adaptability. This result demonstrates that efficient adaptive TA is essential for reducing video freezing and stabilizing quality of experience.

Fig.\ref{fig:freezing_quality} illustrates the average freezing rate as a function of video resolution for the proposed TokenCom framework and baselines, for $U=4$ and $16$ users. At low video resolution, i.e., 360p, the freezing rate is almost zero, but as the video resolution increases, freezing rates increase for all methods. The proposed TokenCom framework maintains a consistently low freezing rate across all resolution levels, outperforming baselines. Moreover, the proposed TokenCom framework demonstrates strong scalability with respect to the number of users, as increasing the number of users from 4 to 16 leads to only a marginal increase in freezing rate. Specifically, in comparison with the conventional H.265 based baseline, the proposed TokenCom framework reduces the freezing rate by roughly 68\%, for high resolution 1080p video, at $U=16$.

\begin{table}[t]
\centering
\caption{\footnotesize Rate-distortion/perception for the considered video tokenizers.}
\label{tokenizers}
\begingroup
\footnotesize
\renewcommand{\arraystretch}{0.5}
\setlength{\tabcolsep}{3pt}
\begin{tabular}{lcccc}
\toprule
\textbf{Tokenizer} & \textbf{PSNR} $\uparrow$ & \textbf{SSIM} $\uparrow$ & \textbf{rFVD} $\downarrow$ & \textbf{bpp} \\
\midrule
Cosmos-0.1-Tokenizer-DV8$\times$16$\times$16 \cite{nvidia} & 25.09 & 0.714 & 241.52 & 0.008\\
Cosmos-0.1-Tokenizer-DV4$\times$8$\times$8 \cite{nvidia}   & 28.81 & 0.818 & 37.36 & 0.063 \\
HEVC medium \cite{zhao2024} & 33.21 & 0.856 & 25.16 & 0.084 \\
BSQ-VAE \cite{zhao2024} & 38.41 & 0.920 & 10.057 & 0.127 \\

\bottomrule
\end{tabular}
\endgroup
\end{table}

\begin{table}[t]
\centering
\caption{\footnotesize Parameter Settings.}
\label{settings}
\begingroup
\footnotesize                    % smaller font (try \scriptsize if needed)
\renewcommand{\arraystretch}{0.4} % tighter rows
\setlength{\tabcolsep}{3pt}       % tighter columns
\begin{tabular}{|l|c|l|c|}
\hline
\textbf{Parameter} & \textbf{Value} & \textbf{Parameter} & \textbf{Value} \\ \hline
$N$ & 32 & episodes & 500 \\ \hline
$R$ & 16 & steps per episode & 100 \\ \hline
$U$ & 4 & batch size & 256 \\ \hline
$B$ & 30 kHz & buffer size & 100\,000 \\ \hline
$P_{\BS}$ & 30 dBm & $\gamma$ & 0.98 \\ \hline
$\kappa$ & 2 & $\epsilon_{\text{start}}$ & 1.0 \\ \hline
$R_{\mathrm{min}}$ & 1 Mbps & $\epsilon_{\text{end}}$ & 0.05 \\ \hline
$[q_{\mathrm{min}},q_{\max}]$ & [18, 36] dB & $\epsilon_{\text{decay}}$ & 0.995 \\ \hline
$[K_{\min},K_{\max}]$ & [0,8] & $\tau$ & 0.005 \\ \hline
$H \times W$ & 1920$\times$1080 &  $\lambda_{\text{pen}}$ & 2.0 \\ \hline
$\rho$ & 24 & $[\alpha, \beta]$ & [2.0, 1.0] \\ \hline
Hidden layers & $256,256$ & Activation & ReLU \\ \hline
Optimizer & Adam & Learning rate & $2{\times}10^{-4}$ \\ \hline
Exploration noise & 0.1 & Network type & Fully connected \\ \hline
\end{tabular}
\endgroup
\end{table}

Fig.~\ref{fig:PSSI_users} plots the average normalized quality score, computed as the mean of the normalized PSNR and SSIM values, which gradually decreases as the number of users increases, due to the reduced per-user transmit power and resource blocks. The proposed TokenCom framework consistently achieves the highest score for various numbers of users and transmit antennas, specifically outperforming the conventional H.265-based baseline. The Fixed-TA exhibits a constant score, as without adaptation capability, the model selects the tokenizer considering the worst channel scenario. Finally, Fig.~\ref{fig:PSSI_power} plots the average normalized quality score, which increases as the BS transmit power increases for different numbers of antennas across various methods. The TokenCom framework consistently outperforms baselines.
\vspace{-0.5cm}
\section{Conclusion}
In this letter, we have proposed a multi-user wireless video TokenCom framework with efficient adaptive tokenizer agreement. We have formulated the corresponding joint tokenizer agreement, resource allocation, and beamforming problem, to simultaneously achieve a high semantic quality and resource efficiency. We have proposed a hybrid DQN-DDPG RL framework to solve the resulting mixed-integer non-convex problem. Simulation results have demonstrated that our proposed framework achieves higher semantic distortion/perception performance compared with baselines at a higher resource efficiency, while showing less frequent video freezing and faster and more stable adaptation.
\vspace{-0.5cm}
\bibliography{references}

@ARTICLE{10981779,
  author={Yang, Shuhan and Shen, Bin and Huang, Xiaoge},
  journal={IEEE Communications Letters}, 
  title={{Optimizing Semantic Spectral Efficiency in Wireless Image Transmission: A PPO-Driven Resource Allocation Scheme}}, 
  year={2025},
  volume={29},
  number={6},
  pages={1466-1470},
  doi={10.1109/LCOMM.2025.3566081}}

@ARTICLE{10122232,
  author={Zhang, Haijun and Wang, Hongyu and Li, Yabo and Long, Keping and Nallanathan, Arumugam},
  journal={IEEE Transactions on Communications}, 
  title={{DRL-Driven Dynamic Resource Allocation for Task-Oriented Semantic Communication}}, 
  year={2023},
  volume={71},
  number={7},
  pages={3992-4004},
  doi={10.1109/TCOMM.2023.3274145}}

@ARTICLE{9832831,
  author={Wang, Yining and Chen, Mingzhe and Luo, Tao and Saad, Walid and Niyato, Dusit and Poor, H. Vincent and Cui, Shuguang},
  journal={IEEE Journal on Selected Areas in Communications}, 
  title={{Performance Optimization for Semantic Communications: An Attention-Based Reinforcement Learning Approach}}, 
  year={2022},
  volume={40},
  number={9},
  pages={2598-2613},
  doi={10.1109/JSAC.2022.3191112}}

@ARTICLE{10845882,
  author={Zhang, Maojun and others},
  journal={IEEE Journal on Selected Areas in Communications}, 
  title={Beamforming Design for Semantic-Bit Coexisting Communication System}, 
  year={2025},
  volume={43},
  number={4},
  pages={1262-1277},
  doi={10.1109/JSAC.2025.3531537}}

@article{nvidia,
  title={Cosmos world foundation model platform for physical {AI}},
  author={Agarwal, Niket and others},
  journal={arXiv preprint arXiv:2501.03575},
  year={2025}
}

@ARTICLE{11175596,
  author={Qiao, Li and Mashhadi, Mahdi Boloursaz and Gao, Zhen and Tafazolli, Rahim and Bennis, Mehdi and Niyato, Dusit},
  journal={IEEE Wireless Communications}, 
  title={Token Communications: A Large Model-Driven Framework for Cross-Modal Context-Aware Semantic Communications}, 
  year={2025},
  volume={32},
  number={5},
  pages={80-88},
  doi={10.1109/MWC.001.2500084}}

@ARTICLE{9955525,
  author={Gündüz, Deniz and others},
  journal={IEEE Journal on Selected Areas in Communications}, 
  title={Beyond Transmitting Bits: Context, Semantics, and Task-Oriented Communications}, 
  year={2023},
  volume={41},
  number={1},
  pages={5-41},
  doi={10.1109/JSAC.2022.3223408}}

@ARTICLE{9955312,
  author={Yang, Wanting and others},
  journal={IEEE Communications Surveys \& Tutorials}, 
  title={Semantic Communications for Future Internet: Fundamentals, Applications, and Challenges}, 
  year={2023},
  volume={25},
  number={1},
  pages={213-250},
  doi={10.1109/COMST.2022.3223224}}

@ARTICLE{10872776,
  author={Xia, Le and Sun, Yao and Liang, Chengsi and Zhang, Lei and Imran, Muhammad Ali and Niyato, Dusit},
  journal={IEEE Wireless Communications}, 
  title={Generative {AI} for Semantic Communication: Architecture, Challenges, and Outlook}, 
  year={2025},
  volume={32},
  number={1},
  pages={132-140},
  doi={10.1109/MWC.003.2300351}}

@article{qiao2025todma,
  title={{ToDMA}: Large Model-Driven Token-Domain Multiple Access for Semantic Communications},
  author={Qiao, Li and B. Mashhadi, Mahdi and Gao, Zhen and Schober, Robert and Gündüz, Deniz},
  journal={arXiv preprint arXiv:2505.10946},
  year={2025},
}

@inproceedings{zhao2024,
 author = {Zhao, Yue and Xiong, Yuanjun and Kr\"{a}henb\"{u}hl, Philipp},
 booktitle = {International Conference on Learning Representations},
 pages = {90844--90868},
 title = {Image and Video Tokenization with Binary Spherical Quantization},
 volume = {2025},
 year = {2025}
}

@ARTICLE{RDP2,
  author={Chen, Jun and Yu, Lei and Wang, Jia and Shi, Wuxian and Ge, Yiqun and Tong, Wen},
  journal={IEEE Journal on Selected Areas in Information Theory}, 
  title={On the Rate-Distortion-Perception Function}, 
  year={2022},
  volume={3},
  number={4},
  pages={664-673},
  doi={10.1109/JSAIT.2022.3231820}}

@article{RDP1,
  title={The Perception-Distortion Tradeoff},
  author={Y. Blau and T. Michaeli},
  journal={IEEE/CVF Conference on Computer Vision and Pattern Recognition},
  year={2018}
}

@ARTICLE{6317156,
  author={Ohm, Jens-Rainer and Sullivan, Gary J. and Schwarz, Heiko and Tan, Thiow Keng and Wiegand, Thomas},
  journal={IEEE Transactions on Circuits and Systems for Video Technology}, 
  title={Comparison of the Coding Efficiency of Video Coding Standards—Including High Efficiency Video Coding ({HEVC})}, 
  year={2012},
  volume={22},
  number={12},
  pages={1669-1684},
  doi={10.1109/TCSVT.2012.2221192}}
\bibliographystyle{IEEEtran}
\end{document}